\documentclass[conference]{IEEEtran}
\usepackage[latin9]{inputenc}
\usepackage{amsmath}
\usepackage{graphicx}
\usepackage{multirow}
\usepackage{multicol}
\usepackage{booktabs}
\usepackage{adjustbox}
\usepackage{xcolor}
\usepackage{xspace}
\usepackage{float}
\usepackage{subcaption}
\usepackage[hidelinks]{hyperref}

\title{Attacking Face Recognition with T-shirts: Database, Vulnerability Assessment and Detection}
\author{\IEEEauthorblockN{M. Ibsen$^1$, C. Rathgeb$^1$, F. Brechtel$^1$, R. Klepp$^1$, K. P\"oppelmann$^1$, A. George$^2$, S. Marcel$^2$, C. Busch$^1$}
\IEEEauthorblockA{1 - Biometrics and Security Research Group \\ Hochschule Darmstadt, Germany \\}
\IEEEauthorblockA{2 -  Biometrics Security and Privacy Group \\ Idiap Research Institute, Switzerland \\}
\textbf{** This work has been submitted to the IEEE for possible publication.} \\ \textbf{Copyright may be transferred without notice, after which this version may no longer be accessible. **}
}

\def\eg{\textit{e.g.}\@\xspace} 
\def\ie{\textit{i.e.}\@\xspace}

\begin{document}
\maketitle

\begin{abstract}
Face recognition systems are widely deployed for biometric authentication. Despite this, it is well-known that, without any safeguards, face recognition systems are highly vulnerable to presentation attacks. In response to this security issue, several promising methods for detecting presentation attacks have been proposed which show high performance on existing benchmarks. However, an ongoing challenge is the generalization of presentation attack detection methods to unseen and new attack types. To this end, we propose a new T-shirt Face Presentation Attack (TFPA) database of 1,608 T-shirt attacks using 100 unique presentation attack instruments. In an extensive evaluation, we show that this type of attack can compromise the security of face recognition systems and that some state-of-the-art attack detection mechanisms trained on popular benchmarks fail to robustly generalize to the new attacks. Further, we propose three new methods for detecting T-shirt attack images, one which relies on the statistical differences between depth maps of bona fide images and T-shirt attacks, an anomaly detection approach trained on features only extracted from bona fide RGB images, and a fusion approach which achieves competitive detection performance. 
\end{abstract}


\section{Introduction}
Biometric systems are systems which recognize individuals based on biological and behavioural characteristics. A specific type of biometric system are face recognition systems which use characteristics of faces to recognize individuals. Face recognition systems are convenient to employ as faces can be captured with commodity hardware at a distance. Additionally, based on advances in deep learning and the availability of large face databases face recognition systems have achieved breakthrough biometric recognition performance on several challenging benchmarks~\cite{Deng-ArcFace-IEEE-CVPR-2019, Meng-MagFace-CVPR-2021}. As a result, face recognition systems are widely used in personal, industrial, and governmental security applications. Despite significant improvements in face recognition over the last decades~\cite{Zhao-FaceRecognSurvey-2003,Abate-2DAnd3DFaceRecognition-2007,LiJain-HandbookOfFaceRecognition-2011}, it has been shown that face recognition systems are vulnerable to presentation attacks (PAs), \eg in~\cite{Chingovska-OnTheEffectivenessOfLocalBinaryPatternInFaceAntgiSpoofing-BIOSIG-2012, Raghavendra-FacePAD-Survey-2017}. The ISO/IEC 30107-1 standard~\cite{ISO-IEC-30107-1-PAD-Framework-160115} defines a PA as a "\textit{presentation to the biometric data capture subsystem with the goal of interfering with the operation of the biometric system}". \par

\begin{figure}[!t]
\centering
\includegraphics[width=\linewidth]{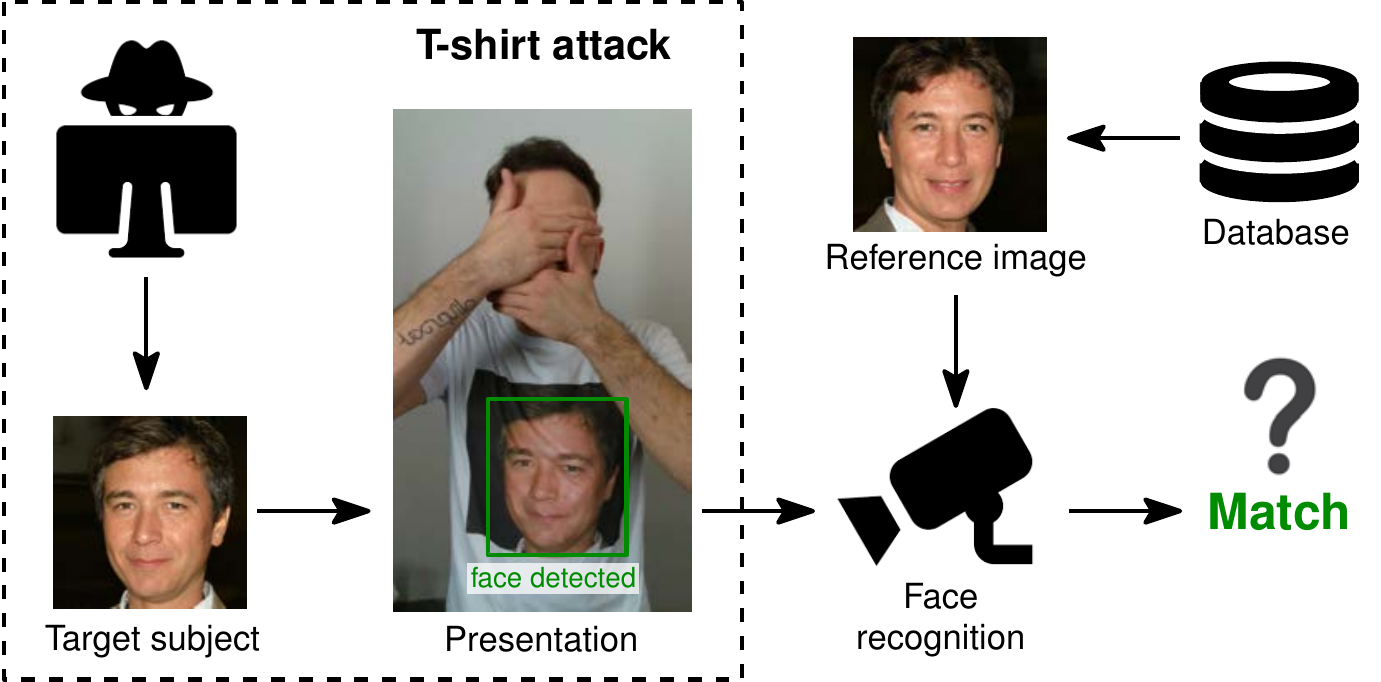}
\caption{Can T-shirts with faces printed on them be used to attack face recognition systems? The T-shirt Presentation Attack Database (TFPA) is introduced along with an extensive benchmark that evaluates the feasibility of the attack, the vulnerability of face recognition systems to the attack, and presentation attack detection.}
\label{fig:tshirt_pa_intro}
\end{figure}

To decrease the vulnerability of PAs, several hardware-based and software-based detection methods have been proposed, along with several datasets containing various types of PAs. Many of the proposed detection methods are trained and evaluated on a specific type of PA or very similar types of PAs. Much work has been conducted on detecting replay and print attacks. However, since then, more sophisticated attacks such as silicon mask~\cite{Raghavendra-CustomSiliconFaceMask-VulnerabilityAndPAD-IWBF-2019} and makeup attacks~\cite{Drozdowski-MakeupPADataset-IWBF-2021} have been proposed, and the focus of many state-of-the-art presentation attack detection (PAD) systems has been on generalizing to unknown attacks. In such a more realistic detection scenario, researchers have reported \emph{significantly higher error} rates compared to evaluating PAD for known or similar attack types~\cite{GonzalezSoler-UnkownAttacksFace-IET-Biometrics-2021}. This underlines that face PAD in real-world scenarios remains an open challenge. \par PAD systems, were initially based on handcrafted features and simple machine learning techniques, but have shifted towards deep learning-based methods. Novel advances include domain adaption~\cite{Wang-PAD-AdvDomainAdaptation-TIFS-2020}, anomaly detection ~\cite{Nikisins-OnEffectivenessOfAnamolyDetectionApproachesAgainstUnseenPresentationAttacksInFaceAntiSpoofing-ICB-2018,Fatemifar-SpoofingAttackDetectionByAnomalyDetection-2019, Ibsen-DifferentialAnomalyDetectionForFacialImages-WIFS-2021}, and new loss functions to better modulate individual channel contribution in multi-stream architectures \cite{George-CMFLForRGBDFaceAntiSpoofing-CVPR-2021}. Despite several advances, generalizability to unknown PAs remains an open problem \cite{Nikisins-OnEffectivenessOfAnamolyDetectionApproachesAgainstUnseenPresentationAttacksInFaceAntiSpoofing-ICB-2018, CostaPazo-FacePADComrehensiveEvaluationGeneralisationProblem-IET-2021}, especially for methods which do not rely on specific hardware-based sensors. As the technology for mitigating against PAs evolves, it is paramount that datasets containing realistic PAs are made available for research. Of special interest are new attack types not yet included in any public dataset and whose impact on face recognition is not publicly known. Consequently, T-shirt face presentation attacks, \ie T-shirts with a human face printed on them, are of particular interest. Said T-shirt attacks have already been identified as being a potential presentation attack instrument (PAI) by border agencies~\cite{EU-Frontex-TechnicalGuideEESRelatedEquipment-2021, BSI-BiometricsPublicSectorApplicationsPart3-2021}. Compared to other PAI species, T-shirt PAs can be worn as part of a regular outfit and, as such, easily hidden, \eg with a cardigan, in semi-controlled environments like automated border control gates. It is an important property for a PAI that it can be easily concealed in case of human interference. Additionally, T-shirt attacks can be fabricated at a relatively low cost making them highly accessible to an attacker. Furthermore, and in contrast to other 2D presentation attacks like videos, it can contain variations of depth across the printed T-shirt face when worn normally by a person which may further hamper its detection by a PAD system. 

The feasibility of a T-shirt PA for concealment, \ie where an attacker tries to hide their own identity and avoid being recognized, was showcased in~\cite{Xu-AdversarialTshirtAttack-2020}. Here, the authors printed adversarial patterns onto T-shirts to avoid automated person detection. However, contrary to this earlier work, the focus of our work is on \textit{impersonation}, \ie where an attacker tries to masquerade a target identity with the goal of being authenticated as that identity. To accomplish this attack, an attacker prints a face image of a target identity onto a T-shirt. The T-shirt is then worn normally and presented by the attacker in-front of the biometric capture device, ideally concealing the real face, \eg using a facial mask, adversarial sticker, or the hands. For the attack to be successful, the face recognition system would detect and process the face on the T-shirt and compare the extracted features to that of the target identity, resulting in a match, \ie a score that surpasses the system's decision threshold. This attack scenario is illustrated in figure~\ref{fig:tshirt_pa_intro}.

This work introduces a database of 100 unique T-shirt PAIs based on synthetically generated face images, each worn by at least two subjects across eight different capturing scenarios and two capturing streams (visible + depth). The database is, to the authors' best knowledge, the first database containing T-shirt PAs for impersonation. The database will be made available\footnote{link will be included upon acceptance of this manuscript}. Additionally, the feasibility of the proposed attacks to circumvent state-of-the-art open-source and commercial face recognition systems is demonstrated in extensive experiments. Furthermore, the vulnerability of existing state-of-the-art PAD algorithms trained on existing benchmark databases to this unknown type of PA is shown and three new methods for detecting T-shirt PAs are proposed.

In summary, this work makes the following contributions:

\begin{itemize}
    \item Fabrication and acquisition of T-shirt attacks for impersonation to compose the new  TFPA database.
    \item An extensive evaluation of the vulnerability of open-source and commercial face recognition systems to the proposed T-shirt attacks, including an analysis of the feasibility of launching the proposed T-shirt attacks.
    \item An evaluation of the generalizability of two state-of-the-art PAD algorithms to the proposed T-shirt attacks.
    \item Proposal and evaluation of three new methods for detecting T-shirt attacks based on (1) statistical differences between depth maps of T-shirt attacks and bona fide images, (2) anomaly detection in the visible spectra, and (3) a fusion approach which combines the scores of (1) and (2).
\end{itemize}

This paper is organized as follows: Section~\ref{sec:related_work} describes the most relevant related work, section~\ref{sec:tshirt_pa} presents the new TFPA database, which is used in section~\ref{sec:experimental_evaluation} to measure the feasibility of the T-shirt PAs and the vulnerability of state-of-the-art face recognition systems to the PAs. Section~\ref {sec:pad_evaluation} evaluates two state-of-the-art open-source PAD algorithms trained on the HQ-WMCA database and three algorithms that do not require training on PAs on the TFPA database. Finally, Section~\ref{sec:discussion} and~\ref{sec:conclusion} discuss and conclude on the findings of this paper.

\section{Related Work}
\label{sec:related_work}
This section presents an overview of face PAs based on clothing and accessories (section~\ref{sec:rw_pas_clothing_accessories}), available multi-channel datasets of face PAs (section~\ref{sec:rw_multichannel_pa_datasets}), the vulnerability studies on face recognition systems against PAs (section~\ref{sec:rw_vulnerability_fr_to_pas}), and a brief overview of recent advances in generalizable face presentation attack detection (section~\ref{sec:rw_face_pad}). The main focus is put on recent software-based PAD techniques.

\begin{table*}[!htb]
    \centering
    \caption{Overview of multi-channel face presentation attack datasets.}
    \begin{tabular}{@{}p{2.2cm}lp{4cm}p{4cm}p{6cm}@{}} \toprule  \textbf{Database} & \textbf{Year} & \textbf{Attack types} & \textbf{Channels} & \textbf{Details}\\ \midrule
    3DMAD~\cite{Erdogmus_3DMAD_BTAS_2013} & 2013 & 3D masks & Visible, depth & 76500 frames of 17 persons. A life-size wearable mask and a paper-cut mask were created for each subject \\  \midrule
    I$^2$BVSD~\cite{Dhamecha-I2BVSD-ICB-2013} & 2013 & 3D facial disguises & Visible, thermal & 75 subjects with disguise variations. For each subject, there is at least one frontal  bona fide image and five frontal disguised images\\  \midrule
    GUC-LiFFAD~\cite{Raghavendra-PADLightFieldCameria-TIP-2015} & 2015 & 2D print (laser, inkjet) and replay & Light-field imagery & 80 subjects with 3 attack categories. Multiple bona fide and attacks per subject. 4826 total samples  \\  \midrule
    MS-Spoof~\cite{Chingovska-FaceRecognitionAcrossImagingSpectrum-Springer-2016} & 2016 & 2D print & Visible, NIR (800nm) & 21 subjects with 5 visible and NIR images across 7 environmental conditions. 3 images in both spectra are selected per client to create black and white print attacks. For each print, 4 spoofing attacks were recorded across 3 lighting conditions in visible and NIR spectra\\  \midrule
    BRSU~\cite{Steiner-DesignActiveMultispectralCameraSystemSkinDetectionFaceVerification-JournalOfSensors-2015,Steiner-ReliableFaceAntispoofingUsingSWIRImaging-ICB-2016} & 2016 & 3D masks and facial disguises & Visible, 4 SWIR bands & Bona fide and presentation attack images at four SWIR bands (935nm, 1060nm, 1300nm, 1550nm) \\  \midrule
    EMSPAD~\cite{Raghavendra-VulnerabilityOfExtendedMultispectralFRTowardsPAs-ISBA-2017} & 2017 &  2D print (laser, inkjet) & 7-band multi-spectral data & 50 subjects across seven spectral bands (425nm, 475nm, 525nm, 570nm, 625nm, 680nm, 930nm) and 2 sessions. Artefacts generated using HP photo print inkjet printer and RICOH laser jet printer \\  \midrule
    MLFP~\cite{Agarwal-FacePresentationAttackWithLatexMasksInMultispectralVideos-CVPRW-2017} & 2017  & 2D paper and 3D latex masks & Visible, NIR, thermal & 1350 videos of 10 subjects (1200 attacks, 150 bona fide). 7 latex masks and 3 paper masks \\  \midrule
    WMCA~\cite{George-WMCA-TIFS-2020} & 2019  & Glasses, fake head, print, replay, rigid mask, flexible mask, paper mask & Visible, depth, infrared, thermal & 72 subjects with 1679 samples (347 bona fide and 1332 attacks) captured over seven sessions\\  \midrule
    HQ-WMCA~\cite{Heusch_TBIOM_2020} & 2020 & Print, replay, rigid mask, paper mask, flexible mask, mannequin, glasses, makeup, tattoo, wig & Visible, depth, thermal, NIR, and SWIR & 51 subjects with 2904 samples (555 bona fide, 2359 attacks) captured over three sessions\\  \midrule
    CASIA-SURF~\cite{Zhang-CASIASURF-TBIOM-2020} & 2020 & Print, cut & Visible, depth, infrared & 21000 videos from 1000 subjects. Different attacks created by printing color face images and cutting certain regions of the face image out, \eg eyes \\ \midrule
    CASIA-SURF CeFA~\cite{Liu-CASIACeFA-WACV-2021} & 2020 & 2D print, replay, 3D masks & Visible, depth, infrared & 23538 videos from 1607 subjects covering 3 ethnic groups and multiple types of 2D and 3D attacks    \\
    \bottomrule
    \end{tabular}
    \label{tab:multichannel_pad_databases_overview}
\end{table*}

\subsection{Face Presentation Attacks using Clothing and Accessories}
\label{sec:rw_pas_clothing_accessories}
Clothing and accessories can be used as attack instruments against face recognition systems. Most known attacks for clothes or accessories focus on printing adversarial patterns onto physical objects which can be worn or attached to the face. In~\cite{Sharif-FrameworkAdversarialExamples-ACM-2019}, the authors illustrate the usage of adversarial generative nets for concealment and impersonation attacks against face recognition systems by printing adversarial patterns onto physical glasses. Although the experiments were limited in size, the results showed successful concealment attacks where in the worst attempt, 81\% of frames in a recorded video were classified as a different identity. For impersonation, an average of 67.58\% of video frames were classified as the target identity, indicating that, in some cases, the proposed adversarial glasses can be used for impersonation. In~\cite{Xu-AdversarialTshirtAttack-2020}, the authors proposed to apply adversarial patterns by using non-rigid objects in the form of T-shirts. The aim of the adversarial patterns was concealment, more specifically, avoiding being detected by a state-of-the-art algorithm. The results showed an attack success rate of 57\% in the physical domain using the YOLOv2 algorithm. In~\cite{Ryu-AdversarialAttacksByAttachingNoiseMarkersAgainstFR-2021-JISA}, white-box attacks were launched in which an attacker has detailed information about a face recognition system. Under this assumption and using only a few subjects, the authors showed that adversarial perturbations in the form of up to 10 noise markers attached to a face can successfully be used for impersonation. \par 
Other works have shown that makeup can reduce face recognition accuracy~\cite{Dantcheva-CanFacialCosmeticsAffectTheMatchingAccuracyOfFaceRecognitionSystems} and even be used for impersonation attacks~\cite{Chen-SpoofingFacesUsingMakeupAnInvestigativeStudy-IEEE-2017}. Especially, it has been found that makeup-induced impersonation attacks of high quality can be used to successfully attack face recognition systems~\cite{Rathgeb-MakeupAttackDetection-IWBF-2020, Drozdowski-MakeupPADataset-IWBF-2021}.

\subsection{Multi-Channel Face Presentation Attack Datasets}
\label{sec:rw_multichannel_pa_datasets}
Today, several datasets containing multi-channel face PAs are available. These datasets cover a range of different attacks, from 2D to 3D attacks. Most databases present a single or a few related attack types; \ie print or replay attacks~\cite{Raghavendra-PADLightFieldCameria-TIP-2015,Chingovska-FaceRecognitionAcrossImagingSpectrum-Springer-2016,Raghavendra-VulnerabilityOfExtendedMultispectralFRTowardsPAs-ISBA-2017, Zhang-CASIASURF-TBIOM-2020,Liu-CASIACeFA-WACV-2021} or 3D mask attacks~\cite{Erdogmus_3DMAD_BTAS_2013,Dhamecha-I2BVSD-ICB-2013,Steiner-ReliableFaceAntispoofingUsingSWIRImaging-ICB-2016,Liu-CASIACeFA-WACV-2021}. An exception is the WMCA dataset and its high-quality variant (HQ-WMCA) which includes several 2D and 3D attacks captured across multiple spectra. Many of the proposed datasets contain images in the visible spectrum plus one or a few additional spectra, \eg depth, thermal, near-infrared (NIR), and short wave infrared (SWIR). An overview of multi-channel face presentation attack datasets is given in table~\ref{tab:multichannel_pad_databases_overview}. It can be observed that none of the existing multi-channel databases includes T-shirt attacks, hence a new database with such attacks can complement existing databases.

\subsection{Vulnerability of Face Recognition to Presentation Attacks}
\label{sec:rw_vulnerability_fr_to_pas}
Multiple works have investigated the effect of different PAs on face recognition performance. In \cite{Chingovska-BiometricEvaluationUnderSpoofingAttacks-TIFS-2014, Raghavendra-FacePAD-Survey-2017,Mohammadi-DeeplyVulnerableAStudy-IET-2017}, different researchers investigated the vulnerability of face recognition systems against print and replay attacks. While these works differ in how the security thresholds of the systems were chosen and used to evaluate the vulnerability of the face recognition systems, they all highlight the feasibility of print and replay attacks. Precisely, success rates in terms of impostor attack presentation match rates (IAPMRs) above 90\% were reported, with a few exceptions most notably the Inter-Session Variability algorithm in~\cite{Mohammadi-DeeplyVulnerableAStudy-IET-2017}. However, as noted by the authors, this system exhibits a high false non-match rate (FNMR) on bona fide images at the same threshold, indicating its limited recognition performance in general. The results indicate that face recognition systems with higher performance on bona fide images are more vulnerable to PAs. In~\cite{Erdogmus-Spoofing2DFRwith3DMasks-BIOSIG-2013, Raghavendra-PADApplicationTo3DFaceMask-ICIP-2014,Erdogmus_3DMAD_BTAS_2013}, the authors reveal the vulnerability of face recognition systems to 3D masks which at the selected threshold show high vulnerability, \ie $>$78\% IAPMR, for most algorithms except for the Inter-Session Variability algorithm evaluated in~\cite{Erdogmus_3DMAD_BTAS_2013} where a IAPMR of 65.7\% is achieved. In a more recent work~\cite{Raghavendra-CustomSiliconFaceMask-VulnerabilityAndPAD-IWBF-2019} and using a new database of custom silicone masks for eight subjects, a less significant vulnerability to 3D masks was reported using two commercial systems. In the worst case corresponding to a false acceptance rate (FAR) of 0.1\%, a IAPMR of 28.20\% was reported. At lower FAR of 0.01\%, the face recognition systems were shown to prevent from 3D mask PAs.\par
An overview of vulnerability studies of different face recognition systems towards different PAs is given in table~\ref{tab:fr_vulnerability_to_pa}. As can be seen, the different face recognition systems report general high vulnerabilities in terms of IAPMR, indicating the need for reliable PAD algorithms. 

\begin{table*}[!htb]
    \centering
    \caption{Overview of the vulnerability of face recognition to presentation attacks. Note: For \cite{Raghavendra-FacePAD-Survey-2017} IAPMR results are reported for the high quality samples and for \cite{Raghavendra-CustomSiliconFaceMask-VulnerabilityAndPAD-IWBF-2019} the IAPMR is reported as an average performance on the iPhone, Samsung S7 and S8 subsets of the CSMad-Mobile dataset.}
    \begin{tabular}{@{}lllllp{4cm}@{}} \toprule  \textbf{Paper} & \textbf{Database} & \textbf{Attack types} & \textbf{Face recognition algorithm} & \textbf{IAPMR \%} & \textbf{Threshold for IAPMR evaluation}\\ \midrule
    \multirow{2}{*}{\cite{Erdogmus-Spoofing2DFRwith3DMasks-BIOSIG-2013}} & \multirow{2}{*}{\cite{Erdogmus-Spoofing2DFRwith3DMasks-BIOSIG-2013}} & \multirow{2}{*}{3D masks} & Gabor graphs & 78.3 & \multirow{2}{*}{Set to EER} \\ 
     & & & Local Gabor Binary Pattern Histogram Sequence & 97.6    \\  \midrule       
     \multirow{4}{*}{\cite{Chingovska-BiometricEvaluationUnderSpoofingAttacks-TIFS-2014}} &  \multirow{4}{*}{Replay-Attack~\cite{Chingovska-OnTheEffectivenessOfLocalBinaryPatternInFaceAntgiSpoofing-BIOSIG-2012}} &  \multirow{4}{*}{print, replay} & Gaussian Mixture Model &   91.5 &  \multirow{4}{*}{Set to EER}\\
                &                              &                         &  Local Gabor Binary Pattern Histogram Sequence    & 88.5   \\
                &                                  &                         & Gabor Jet & 95.0   \\
                &                                  &                         & Inter-Session Variability & 92.6  \\ \midrule
        \cite{Erdogmus_3DMAD_BTAS_2013} & 3DMAD~\cite{Erdogmus_3DMAD_BTAS_2013}   & 3D masks & Inter-Session Variability & 65.7 & Set to EER \\ \midrule
    \cite{Raghavendra-PADApplicationTo3DFaceMask-ICIP-2014} &          3DMAD~\cite{Erdogmus_3DMAD_BTAS_2013}             &  3D masks & Sparse Representation Classifier & 84.1 & Set to EER\\ \midrule
    \cite{Raghavendra-FacePAD-Survey-2017} &  CASIA-MFSD & print, replay & VeriLook &  100 & Set to FMR=0.01\% \\ \midrule
    \multirow{5}{*}{\cite{Mohammadi-DeeplyVulnerableAStudy-IET-2017}} &  & \multirow{5}{*}{print, replay} & VGG-Face & 92.7 & \multirow{5}{*}{Set to FMR=0.1\%}\\
    & & & LightCNN & 99.6  \\
    & Combination of & & FaceNet & 99.8  \\
    & \cite{Chingovska-OnTheEffectivenessOfLocalBinaryPatternInFaceAntgiSpoofing-BIOSIG-2012,CostaPazo-ReplayMobile-BIOSIG-2016,Wen-MSUMFSD-TIFS-2015}& & ROC-SDK & 84.5 \\
    & & &  Inter-Session Variability & 36.1\\ \midrule
    \multirow{2}{*}{\cite{Raghavendra-CustomSiliconFaceMask-VulnerabilityAndPAD-IWBF-2019}} & \multirow{2}{*}{CSMad-Mobile} &  \multirow{2}{*}{3D mask} &  Neurotech &  16.84 &  \multirow{2}{*}{Set to FAR = 0.1\%}\\ 
    & &  & Cognitec &  14.28 & \\ 
    \bottomrule
    \end{tabular}
    \label{tab:fr_vulnerability_to_pa}
\end{table*}

\subsection{Face Presentation Attack Detection}
\label{sec:rw_face_pad}
Several face PAD methods have been proposed in recent years, including hardware and software-based approaches. Recent advances have mostly focused on deep learning-based approaches. Further, anomaly detection and domain adoption have also shown promising results. \par In anomaly detection, a classifier is usually trained on bona fide data. In~\cite{George-OneClassPADUsingMCCNN-TIFS-2021}, the authors trained a multi-channel convolution neural network to learn compact face embeddings of bona fide images, which were used in conjunction with a one-class Gaussian mixture model (GMM). The approach showed good results on the \textit{grandtest} protocol of the multi-channel WMCA database as well as when evaluated for generalisability using only the RGB stream. However, the results show that there is still room for improvement. Specifically, the authors reported a detection equal-error-rate (D-EER) of 12\% on average when the method was evaluated using a leave-one-out protocol on the SiW-M dataset. Recently in~\cite{Ibsen-DifferentialAnomalyDetectionForFacialImages-WIFS-2021}, an identity-aware anomaly detection approach was presented where deep face embeddings extracted from a trusted and suspected images were combined and used to train a one-class classifier for detecting anomalies. Trained on only bona fide data, the approach showed high generalizability to both digital and physical attacks, including silicon mask and makeup attacks. \par For domain-adaption, the main idea is to overcome poor cross-dataset performance by transferring from a source domain to a target domain. In~\cite{Wang-PAD-AdvDomainAdaptation-TIFS-2020}, an unsupervised domain adaption approach was introduced, which at the time, showed superior performance compared to other approaches for several popular benchmarks in cross-database evaluations. \par
Another interesting approach was recently proposed in~\cite{GonzalezSoler-UnkownAttacksFace-IET-Biometrics-2021} where the idea was to explore semantic information of known attacks and bona fide samples and use the information to learn to detect PAs. The approach achieved a D-EER of 18.24\% when evaluated in a cross-database setup of three popular benchmarks. Other prominent approaches include the use of vision transformers for PAD~\cite{George-VIT-IJCB-2021}, which showed state-of-the-art performance on challenging datasets such as the leave-one-out protocol on the SiW-M dataset. More precisely, in~\cite{George-VIT-IJCB-2021} an average D-EER of 6.72\% was achieved. The authors also proposed a cross modal focal loss (CMFL)~\cite{George-CMFLForRGBDFaceAntiSpoofing-CVPR-2021} for supervising individual channels in a multi-channel and multi-head architecture. Although trained on multi-channel data, the separate heads allow using the architecture even in cases where a specific stream is not available. Using RGB and depth streams, the authors demonstrated the effectiveness of the approach. However, in a cross-database evaluation between WMCA and HQ-WMCA, the performance was rather poor, \ie a half total error rate (HTER) of 29.1\% when trained on WMCA and tested on HQ-WMCA. Slightly better results were achieved when trained on HQ-WMCA and tested on WMCA, achieving an HTER of 18.2\%. 

\section{T-shirt Face Presentation Attack Database}
\label{sec:tshirt_pa}
\begin{figure}[!t]
\centering
\begin{subfigure}[t]{.48\columnwidth}
\centering
\includegraphics[width=.48\columnwidth]{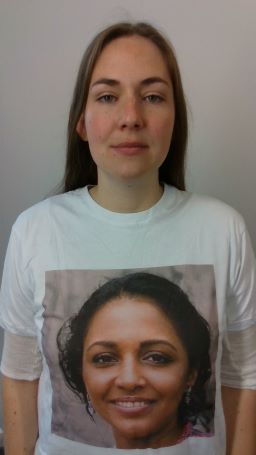}
\includegraphics[width=.48\columnwidth]{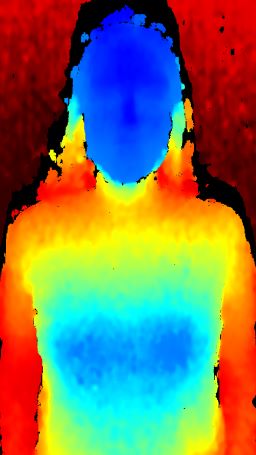}
\end{subfigure}\hspace{6pt}%
\begin{subfigure}[t]{.48\columnwidth}
\centering
\includegraphics[width=.48\columnwidth]{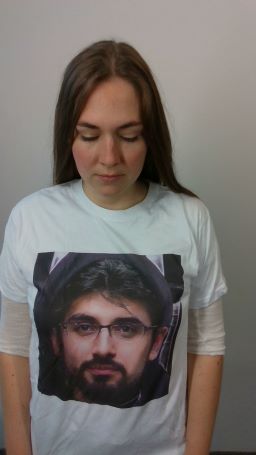}
\includegraphics[width=.48\columnwidth]{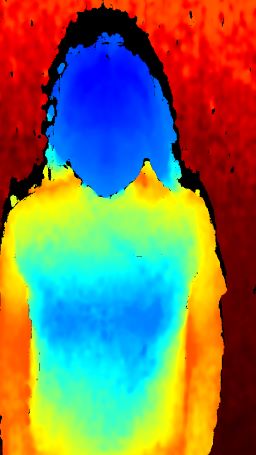}
\end{subfigure}

\vspace{3pt}
\begin{subfigure}[t]{.48\columnwidth}
\centering
\includegraphics[width=.48\columnwidth]{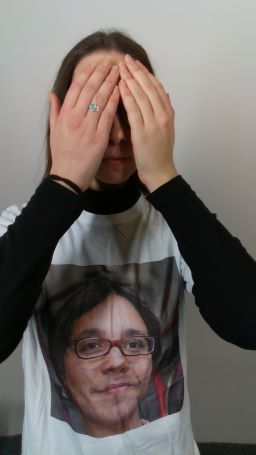}
\includegraphics[width=.48\columnwidth]{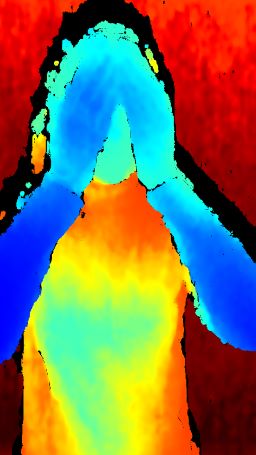}
\end{subfigure}\hspace{6pt}%
\begin{subfigure}[t]{.48\columnwidth}
\centering
\includegraphics[width=.48\columnwidth]{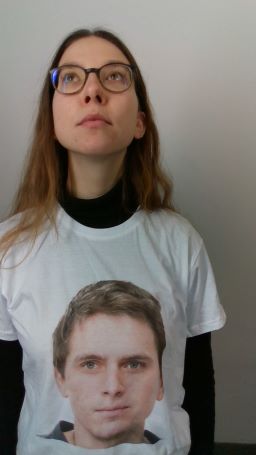}
\includegraphics[width=.48\columnwidth]{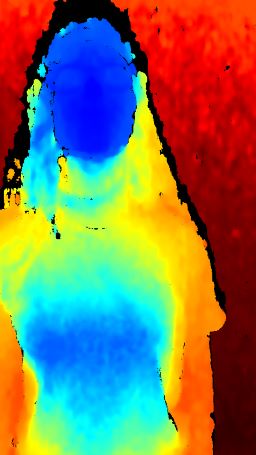}
\end{subfigure}
\caption{Example attacks from the proposed T-shirt presentation attack database. Based on synthetic T-shirt face images.}
\label{fig:image_pairs_examples}
\end{figure}

The multi-channel T-shirt Face Presentation Attack (TFPA) Database consists of 1,608 T-shirt PAs captured using 100 unique T-shirts with face images printed on them. A total of eight subjects participated in the data collection and each T-shirt was worn by at least two different subjects. For each T-shirt worn by a subject, the subject was asked to take up eight different poses resulting in eight different capturing scenarios per T-shirt worn by a subject. Both visible (RGB) and depth data were captured. An overview of the TFPA database is given in table~\ref{tab:db_description} which describes the different properties of the database. Examples of RGB and depth images for TFPA can be seen in figure~\ref{fig:image_pairs_examples}.

\begin{table}[!htb]
    \centering
    \caption{An overview of the proposed T-shirt face presentation attack (TFPA) database.}
    \begin{tabular}{@{}ll@{}} \toprule
    \textbf{Property} & \textbf{Description}\\ \midrule
    Attack type: & Impersonation attack \\
    Attack creation: & Synthetic face printed on T-shirt \\
    Face generation methods: & StyleGAN \& InterFaceGAN \\
    No. of spectrums: & 2 (visible, depth)   \\
    No. of attacks: & 1608 \\
    No. of PAIs: & 100 \\
    No. of subjects: & Real (8), generated (100) \\
    Capturing device: & Intel RealSense Depth Camera D435 \\
    Environment: & Controlled indoor with white background \\
    \bottomrule
    \end{tabular}
    \label{tab:db_description}
\end{table}

\subsection{Creation of T-shirt attacks}
\begin{figure}[!tb]
\centering
\begin{subfigure}{0.19\linewidth}
    \centering %
  \includegraphics[width=\textwidth]{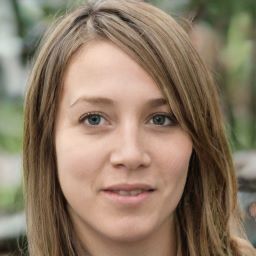}
\end{subfigure}
\begin{subfigure}{0.19\linewidth}
    \centering %
  \includegraphics[width=\textwidth]{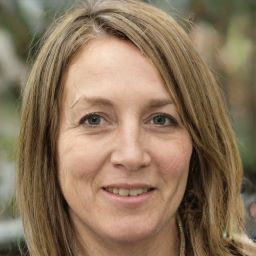}
\end{subfigure}
\begin{subfigure}{0.19\linewidth}
    \centering %
  \includegraphics[width=\textwidth]{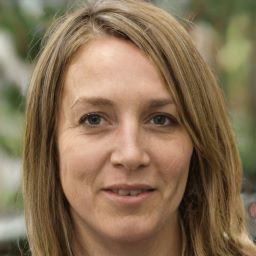}
\end{subfigure}
\begin{subfigure}{0.19\linewidth}
    \centering %
  \includegraphics[width=\textwidth]{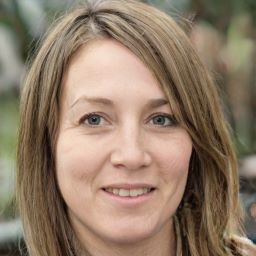}
\end{subfigure}
\begin{subfigure}{0.19\linewidth}
    \centering %
  \includegraphics[width=\textwidth]{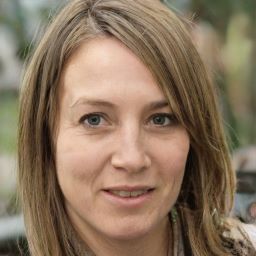}
\end{subfigure}

\vspace{3pt}
\begin{subfigure}{0.19\linewidth}
    \centering %
  \includegraphics[width=\textwidth]{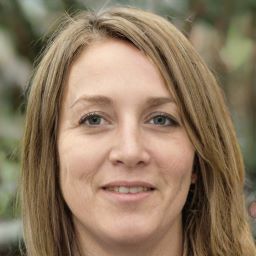}
\end{subfigure}
\begin{subfigure}{0.19\linewidth}
    \centering %
  \includegraphics[width=\textwidth]{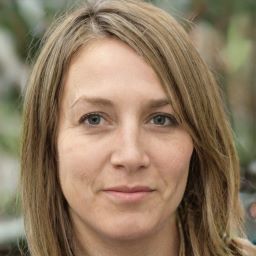}
\end{subfigure}
\begin{subfigure}{0.19\linewidth}
    \centering %
  \includegraphics[width=\textwidth]{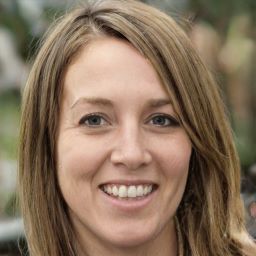}
\end{subfigure}
\begin{subfigure}{0.19\linewidth}
    \centering %
  \includegraphics[width=\textwidth]{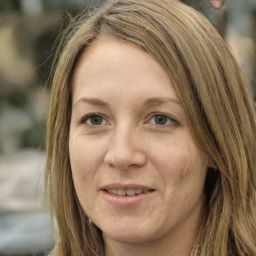}
\end{subfigure}
\begin{subfigure}{0.19\linewidth}
    \centering %
  \includegraphics[width=\textwidth]{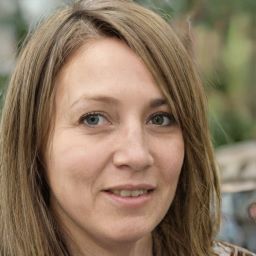}
\end{subfigure}
\caption{Example of generated intra-class variations for a single synthetical identity. }
\label{fig:subject_intra_class_variations}
\end{figure}

To create the T-shirt attacks, 150 synthetic images were randomly generated using StyleGAN~\cite{Karras-Stylegan1-CVPR-2019} and subsequently filtered to 100 images to have a good distribution with respect to sex, skin tone, and age. Then variations of each subject were created by changing the age, yaw, smile, pitch, and illumination of the synthetically created images using InterFaceGAN~\cite{Shen-InterFaceGAN-TPAMI-2022}. We used the default boundaries provided in the InterFaceGAN repository for age, yaw, and smile. The boundaries for pitch and illumination were trained manually on a small labelled dataset of StyleGAN images. An example of the 10 different intra-class variations for a subject is given in figure~\ref{fig:subject_intra_class_variations}. For each subject, a single neutral image was selected and printed onto a T-shirt. The 10 different intra-class variations of a subject are used as reference images when doing the face recognition evaluation in section~\ref{sec:fr_evaluation}.\par The face images were printed on medium-sized white T-shirts to best maintain the colour of the face images and to make the face images cover most of the T-shirts without stretching the facial images. 

\subsection{Data Capture Setup}
For data acquisition, the Intel RealSense Depth Camera D435 was used. The Intel RealSense Depth Camera D435 is a RGB-D sensor consisting of a pair of depth sensors, a RGB sensor and an infrared projector. It can capture depth images with an operational range of up to 10 meters and a depth frame rate of up to 90 frames per second. It is customizable using the Intel RealSense SDK 2.0. All captured RGB and depth images have resolution 1280$\times$720.

The data acquisition occurred in a bright room with two studio light fixtures to ensure good illumination exposure. A plain white wall was used as the background, and the D435 RGB-D sensor was mounted on a stable tripod. The capture subject was positioned at a distance so the camera could capture from the head down to the waistline.

\subsection{Data collection Procedure}
\begin{figure}[!tb]
\begin{subfigure}{0.24\linewidth}
    \centering %
  \includegraphics[width=\textwidth]{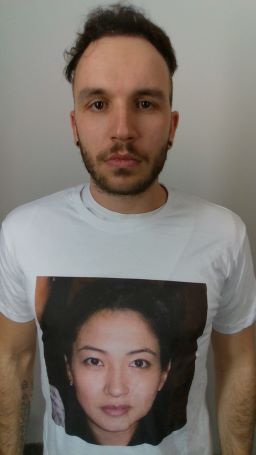}
  \caption{normal}
\end{subfigure}
\begin{subfigure}{0.24\linewidth}
    \centering %
  \includegraphics[width=\textwidth]{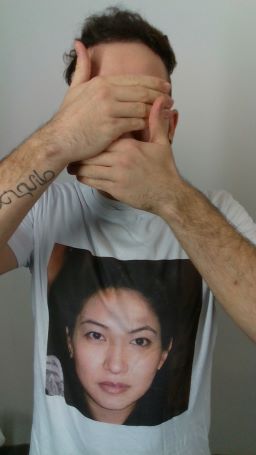}
  \caption{face covered}
\end{subfigure}
\begin{subfigure}{0.24\linewidth}
    \centering %
  \includegraphics[width=\textwidth]{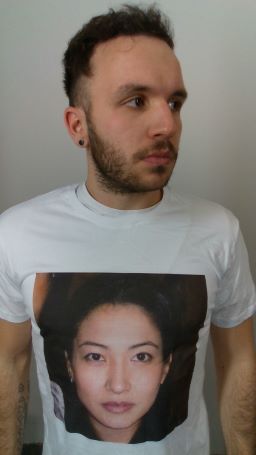}
  \caption{look left}
\end{subfigure}
\begin{subfigure}{0.24\linewidth}
    \centering %
  \includegraphics[width=\textwidth]{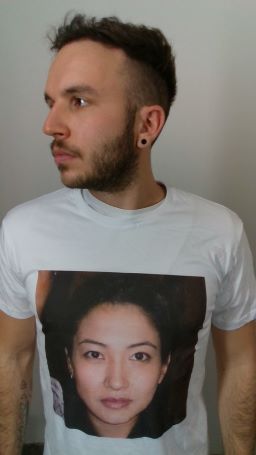}
  \caption{look right}
\end{subfigure}

\vspace{3pt}
\begin{subfigure}{0.24\linewidth}
    \centering %
  \includegraphics[width=\textwidth]{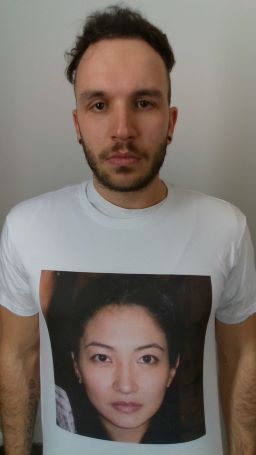}
  \caption{stretch shirt}
\end{subfigure}
\begin{subfigure}{0.24\linewidth}
    \centering %
  \includegraphics[width=\textwidth]{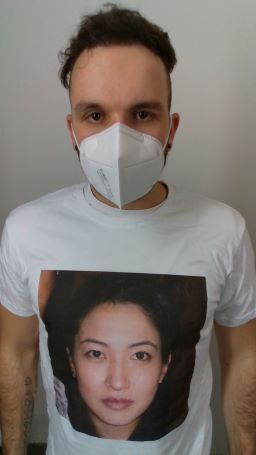}
  \caption{facial mask}
\end{subfigure}
\begin{subfigure}{0.24\linewidth}
    \centering %
  \includegraphics[width=\textwidth]{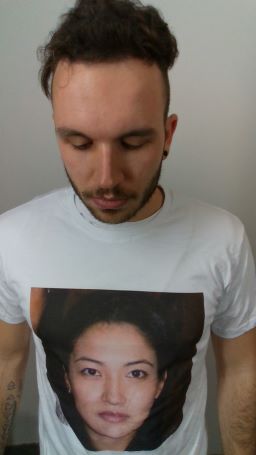}
  \caption{look down}
\end{subfigure}
\begin{subfigure}{0.24\linewidth}
    \centering %
  \includegraphics[width=\textwidth]{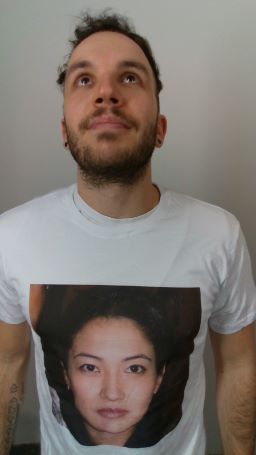}
  \caption{look up}
\end{subfigure}
\caption{The eight different capturing scenarios. The real face of the capture subject is fully covered in (b) and partially covered by the face mask in (f).}
\label{fig:capturing_scenarios}
\end{figure}

The capturing subjects were asked to wear different T-shirts across the eight capturing scenarios depicted in figure~\ref{fig:capturing_scenarios}, resulting in eight RGB and depth images per subject and worn T-shirt. The capturing scenarios has been designed to accommodate different types of presentations to the biometric capturing device simulating variations in presentations for both the real and the T-shirt face. For the real face, the variations include occlusions in form of face masks and hands as well as pose variations. For the T-shirt face, the different poses and movements of the subject cause the facial image to sometimes appear as being on a flat surface and other times appear with varying degree of depth deformations. The different capturing scenarios are important as it will allow analysing in which cases it is feasible to launch this type of PA.

\section{Vulnerability Evaluation}
\label{sec:experimental_evaluation}
In this section, we measure the vulnerability of state-of-the-art face recognition systems to T-shirt PAs. To do so, we divide the experimental evaluation into two parts: (1) analysis on the feasibility of launching the attack and (2) an analysis on the success chance of the T-shirt attacks in terms of IAPMR. The reason for evaluating (1) and (2) separately is that compared to other types of PAs where the PAI covers the real face, T-shirt PAs present an additional challenge for the attacker if the T-shirt is worn normally. In this case, to launch the attack, attackers must take special care to hide their real face and ensure that, at best, only the face on the T-shirt is detected. Hence, we estimate the success rate of launching the T-shirt attack using three face detection algorithms. In the second part, we assume that the face on the T-shirt has been detected by a face recognition system and measure how often it is successfully reaching a match decision in a comparison trial against a stored reference image.

An overview of the algorithms used for the different stages of the evaluation is given in table~\ref{tab:modules_algortihms}. COTS refer to the commercial-of-the-shelf face recognition system. 

\begin{table}[!htb]
    \centering
    \caption{An overview of the different algorithms used in the different evaluation stages.}
    \begin{tabular}{@{}lp{5cm}@{}} \toprule \textbf{Module} & \textbf{Algorithms}   \\ \midrule 
    Face detection & dlib~\cite{King-MachineLearningToolkit-2009}, MTCNN~\cite{Zhang-JointFaceDetectionAndAllignmentUsingMultitaskCascadedConvolutionalNetworks-2016}, RetinaFace~\cite{deng-retinaface-cvpr-2020}\\
    Face recognition &  ArcFace~\cite{Deng-ArcFace-IEEE-CVPR-2019}, COTS \\
    \bottomrule
    \end{tabular}
    \label{tab:modules_algortihms}
\end{table}

\subsection{Face Detection}

\subsubsection{Experiments}

To properly evaluate the effect of the T-shirt attacks on face recognition systems, we must analyze how face detection algorithms deal with faces printed on T-shirts. If the faces on the T-shirts are not properly detected or if the real face is detected with much higher confidence, the attack is likely filtered out in the pre-processing phase of the face recognition system and hence won't be able to spoof the face recognition system. To evaluate this, three open-source algorithms were used, namely RetinaFace~\cite{deng-retinaface-cvpr-2020}, MTCNN~\cite{Zhang-JointFaceDetectionAndAllignmentUsingMultitaskCascadedConvolutionalNetworks-2016}, and dlib~\cite{King-MachineLearningToolkit-2009}. 

\begin{figure}[!htb]
    \centering %
\begin{subfigure}[t]{0.3\linewidth}
  \includegraphics[width=\linewidth]{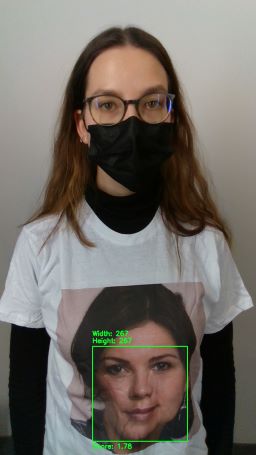}
\end{subfigure} 
\begin{subfigure}[t]{0.3\linewidth}
  \includegraphics[width=\linewidth]{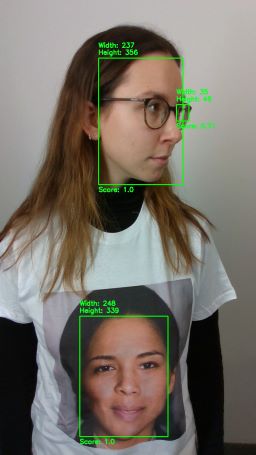}
\end{subfigure}
\begin{subfigure}[t]{0.3\linewidth}
  \includegraphics[width=\linewidth]{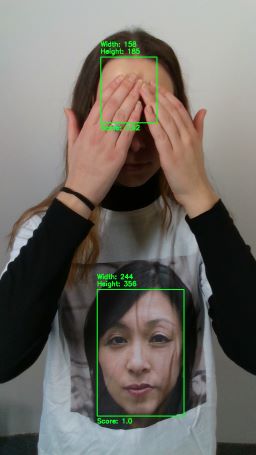}
\end{subfigure}
\caption{Examples of detection errors}
\label{fig:face_detection_errors}
\end{figure}

Face detection was then performed for each algorithm on each presentation attack. This resulted in some detection errors where a region of interest did not include a real face or where the real face or the T-shirt face was not detected. Figure~\ref{fig:face_detection_errors} gives examples of such detection errors. 

\begin{figure}[!htb]
    \begin{subfigure}[t]{0.5\linewidth}
    \centering
        \includegraphics[width=\linewidth]{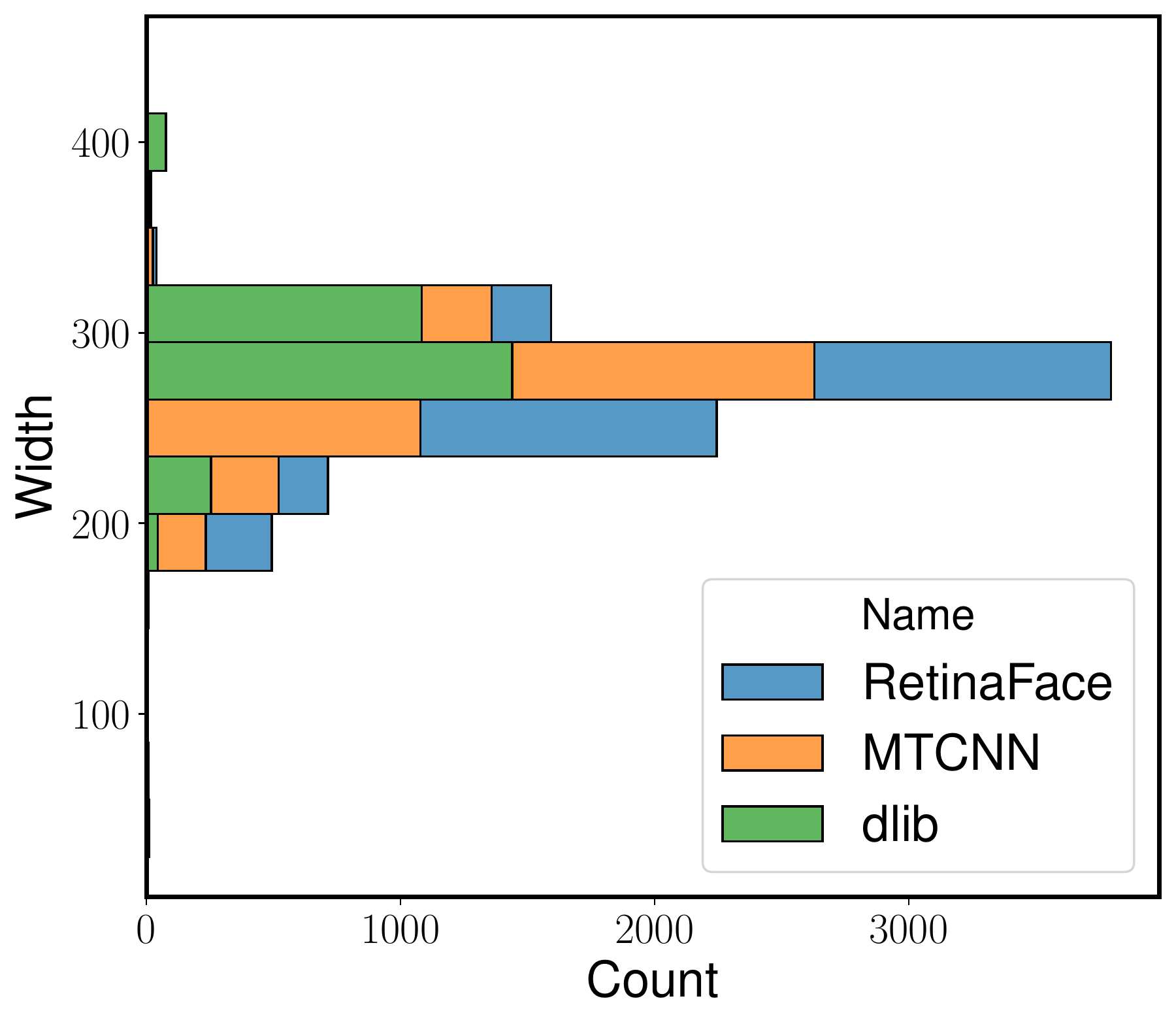}
        \caption{}
        \label{fig:depthmap}
    \end{subfigure}%
    \begin{subfigure}[t]{0.5\linewidth}
    \centering
        \includegraphics[width=\linewidth]{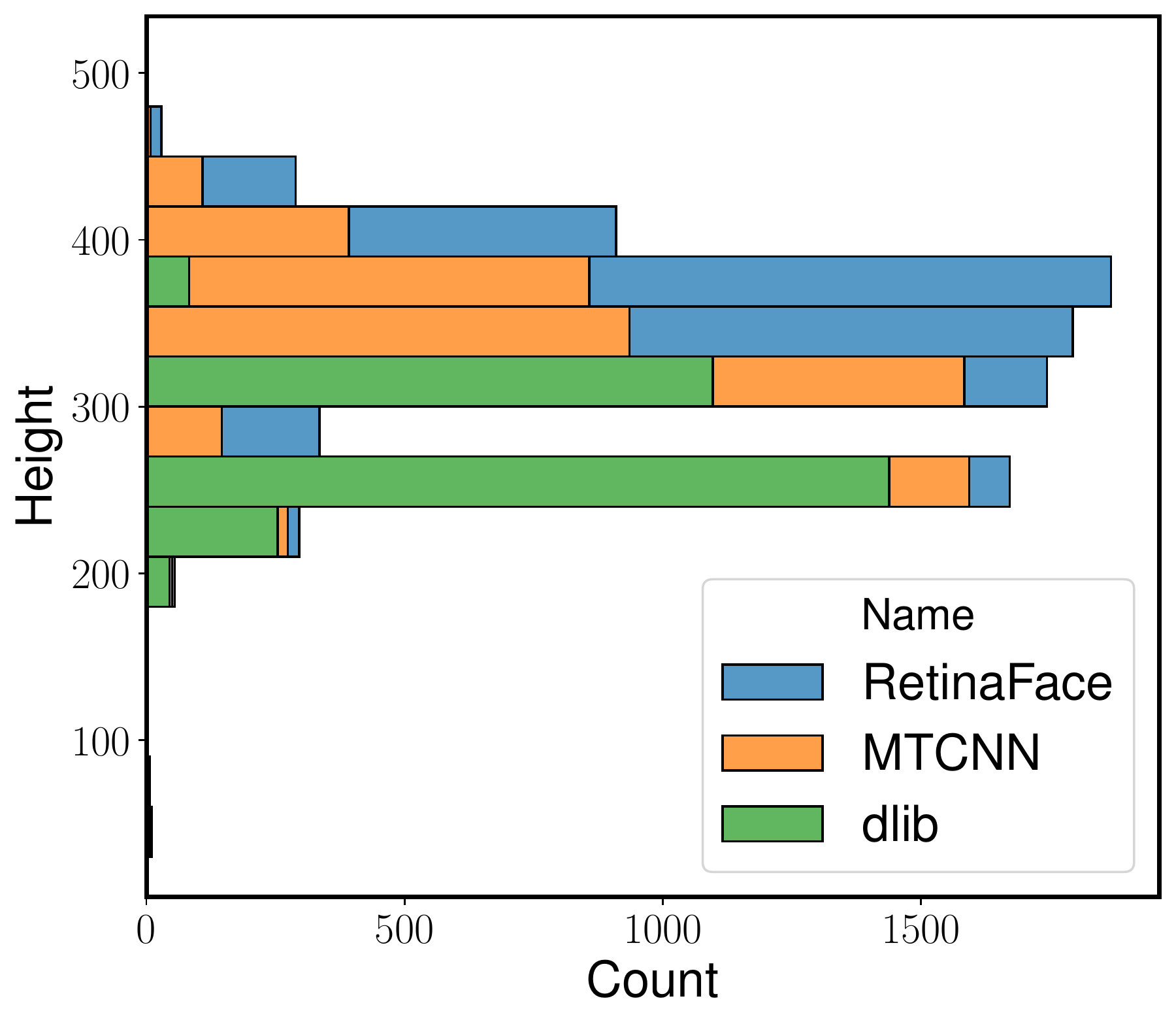}
        \caption{}
    \end{subfigure}
    \caption{Width (a) and height (b) distributions of detected faces.}
        \label{fig:dimensions_detected_regions}
    
\end{figure}

By looking at the width and height of each detected region of interest for the three different algorithms in figure~\ref{fig:dimensions_detected_regions}, we can observe several outliers, which indicate several faces which have been inaccurately detected. As we are only interested in areas with a face, these incorrect detections are removed from the results. 

\subsubsection{Metrics}
To measure the detection capability, we record for each algorithm and a detected region the confidence score of the algorithm, which is a measurement of the likelihood that the region contains a face. We report the average confidence score for real and T-shirt faces for the three detection algorithms and eight capturing scenarios. A Min-Max normalization is performed over the confidence scores individually for each detection algorithm after filtering away the small faces. Additionally, for each scenario and detection algorithm, we estimate how often the T-shirt face or real face are correctly detected. Specifically, we denote the success rate of a detection algorithm as the estimated proportion of faces that were correctly detected. We report this separately for real and T-shirt faces.

\subsubsection{Results}

By analyzing the remaining regions of interest, we can estimate how often a T-shirt face and the real face were successfully detected. Additionally, we can measure the average detection score of each algorithm for the real and T-shirt faces. The results for each capturing scenario and algorithm are summarized in table~\ref{tab:detection_performance_tshirt_attacks}. The results show that the T-shirt faces are successfully detected in almost all cases with an average estimated detection rate for the three algorithms $>99\%$ across all eight poses. The results also show that it is possible to prevent the real face from being detected by, \eg covering the face with the hands or, for some algorithms, increase the chances of launching the attack by wearing a face mask or tilting the face. The results indicate that it is possible to launch the T-shirt attacks in semi-supervised or unsupervised scenarios. However, to increase the success rate some effort should be taken to conceal the real face. 

\begin{table*}[!htpb]
    \centering
\caption{Detection accuracy and average detection scores across algorithms and capturing scenarios for T-shirt and real faces.}
\begin{adjustbox}{max width=\textwidth}
   \begin{tabular}{@{\extracolsep{2pt}}llrrrrrrrr@{}} \toprule 
       & & \multicolumn{2}{c}{\textbf{dlib}}  &
      \multicolumn{2}{c}{\textbf{MTCNN}} & \multicolumn{2}{c}{\textbf{RetinaFace}}  &
      \multicolumn{1}{c}{\textbf{Avg.}} \\
      \cmidrule{3-4} \cmidrule{5-6} \cmidrule{7-8} \cmidrule{9-9}
Scenario & Face type  &  \multicolumn{1}{c}{Success \%} & \multicolumn{1}{c}{Avg. score}  & \multicolumn{1}{c}{Success. \%} & \multicolumn{1}{c}{Avg. score} & \multicolumn{1}{c}{Success \%} & \multicolumn{1}{c}{Avg. score} & \multicolumn{1}{c}{Success \%} \\
\midrule
   \multirow{2}{*}{Normal} 
   & real & 100 & 0.60 & 100 & 1.00 & 100 & 0.98 & 100  \\ 
   & T-shirt & 100 & 0.56 & 100 & 1.00 & 100 & 0.98 & 100  \\ 
   \midrule
      \multirow{2}{*}{Face covered} 
   & real & 0.50 & 0.01 & 10.45 & 0.41 & 11.94 & 0.67 & 7.63  \\ 
   & T-shirt & 98.01 & 0.50 & 99.50 & 1.00 & 100 & 0.97 & 99.17  \\    \midrule
      \multirow{2}{*}{Look left} 
   & real & 88.56 & 0.30 & 100 & 1.00 & 100 & 0.98 & 96.19  \\ 
   & T-shirt & 100 & 0.52 & 100 & 1.00 & 100 & 0.98 & 100  \\ 
   \midrule
     \multirow{2}{*}{Look right} 
   & real & 95.02 & 0.43 & 100 & 1.00 & 100 & 0.98 & 98.34  \\ 
   & T-shirt & 100 & 0.53 & 100 & 1.00 & 100 & 0.98 & 100  \\ 
   \midrule
      \multirow{2}{*}{Stretch T-shirt} 
   & real & 100 & 0.59 & 100 & 1.00 & 100 & 0.97 & 100  \\ 
   & T-shirt & 100 & 0.60 & 100 & 1.00 & 100 & 0.98 & 100  \\ 
   \midrule
      \multirow{2}{*}{Facial mask} 
   & real & 92.04 & 0.17 & 100 & 1.00 & 100 & 0.98 & 97.35  \\ 
   & T-shirt & 99.50 & 0.56 & 100 & 1.00 & 100 & 0.98 & 99.83  \\ 
   \midrule
      \multirow{2}{*}{Look down} 
   & real  & 88.56 & 0.34 & 100 & 1.00 & 100 & 0.98 & 96.19  \\ 
   & T-shirt & 100 & 0.56 & 100 & 1.00 & 100 & 0.98 & 100  \\ 
   \midrule
      \multirow{2}{*}{Look up} 
   & real & 89.55 & 0.31 & 99 & 0.99 & 100 & 0.99 & 96.18  \\ 
   & T-shirt & 100 & 0.55 & 100 & 1.00 & 100 & 0.98 & 100  \\ 
        \bottomrule
    \end{tabular}
\label{tab:detection_performance_tshirt_attacks}
\end{adjustbox}{}
\end{table*}

\subsection{Face Recognition}
This section aims to evaluate the spoofing capabilities of the T-shirt PAs on state-of-the-art open-source and commercial face recognition systems.

\subsubsection{Experiments}
\begin{figure}[!tb]
\begin{subfigure}{0.19\linewidth}
    \centering %
  \includegraphics[width=\textwidth]{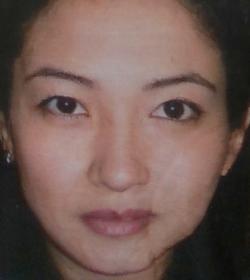}
\end{subfigure}
\begin{subfigure}{0.19\linewidth}
    \centering %
  \includegraphics[width=\textwidth]{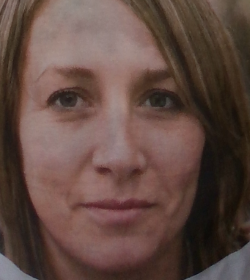}
\end{subfigure}
\begin{subfigure}{0.19\linewidth}
    \centering %
  \includegraphics[width=\textwidth]{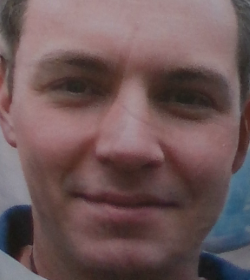}
\end{subfigure}
\begin{subfigure}{0.19\linewidth}
    \centering %
  \includegraphics[width=\textwidth]{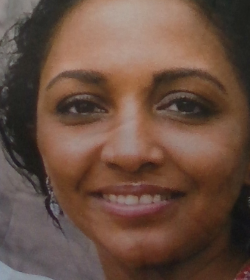}
\end{subfigure}
\begin{subfigure}{0.19\linewidth}
    \centering %
  \includegraphics[width=\textwidth]{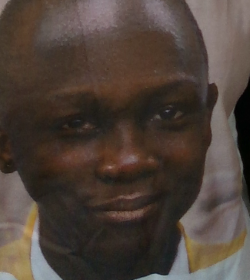}
\end{subfigure}

\vspace{3pt}
\begin{subfigure}{0.19\linewidth}
    \centering %
  \includegraphics[width=\textwidth]{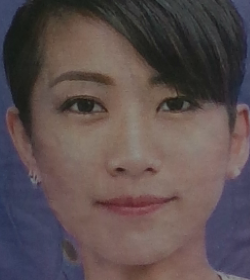}
\end{subfigure}
\begin{subfigure}{0.19\linewidth}
    \centering %
  \includegraphics[width=\textwidth]{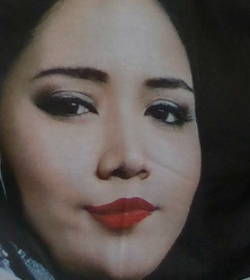}
\end{subfigure}
\begin{subfigure}{0.19\linewidth}
    \centering %
  \includegraphics[width=\textwidth]{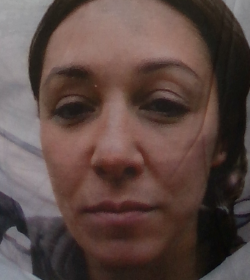}
\end{subfigure}
\begin{subfigure}{0.19\linewidth}
    \centering %
  \includegraphics[width=\textwidth]{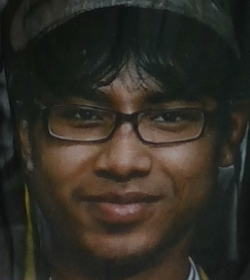}
\end{subfigure}
\begin{subfigure}{0.19\linewidth}
    \centering %
  \includegraphics[width=\textwidth]{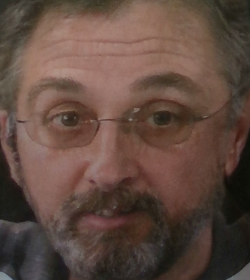}
\end{subfigure}
\caption{Example of cropped out T-shirt faces.}
\label{fig:tshirt_crops}
\end{figure}

Considering the results of the face detection, it was shown that several state-of-the-art algorithms will almost always detect faces on the T-shirt with high confidence and that in some scenarios the real face can be hidden. Therefore, in this subsequent evaluation, the cropped T-shirt faces are fed to the feature extraction of the face recognition systems. This simulates a realistic face recognition pipeline where a face is first detected and then separated from the background before being given as input to the feature extraction and comparison algorithm. Doing this allows us to evaluate if the face on the T-shirt successfully reached a match against a stored reference image. Examples of cropped T-shirt faces can be seen in figure~\ref{fig:tshirt_crops}.

The success of the attack will be evaluated across two scenarios. In the first case, it is assumed that the T-shirt face is compared against a corresponding digital face image, \ie the digital version of the face that is printed on the T-shirt; this would correspond to the case where the attacker has knowledge and access to the exact image stored as a reference image, for instance by reading out the face image from a stolen passport. In the second case, it is assumed that the attacker only has access to different face images of the subject whose face is printed on the T-shirt, \ie that the attacker knows the identity but not the exact image stored in the reference database. We refer to these two scenarios as the \emph{known} and \emph{unknown} scenarios, respectively. To this end, one open-source face recognition (ArcFace~\cite{Deng-ArcFace-IEEE-CVPR-2019}) and one commercial system are used. Additionally, a subset of constrained bona fide images from the FRGCv2~\cite{Phillips-FRGC-2005} dataset is used in order to compare obtained scores with those of bona fide mated and non-mated comparison trials corresponding to semi-controlled access control. An overview of the comparisons for the different datasets is given in table~\ref{tab:FR_comparisons_overview}.

\begin{table}[!htb]
    \centering
    \caption{Overview over the different mated and non-mated comparison trials used in the face recognition evaluation (section~\ref{sec:fr_evaluation}).} 
    \begin{tabular}{@{}lll@{}} \toprule \multirow{2}{*}{\textbf{Scenario}} &  \multicolumn{2}{c}{\textbf{Nr. of comparisons}} \\ \cmidrule{2-3}
    & ArcFace & COTS \\ \midrule
    T-shirt mated (known) & 1,608 & 1.605 \\
    T-shirt mated (unknown) & 14,472 & 14,445 \\
    FRGCv2 mated & 3,292 & 3,292 \\
    FRGCv2 non-mated & 282,492 & 282,492  \\
    \bottomrule
    \end{tabular}
    \label{tab:FR_comparisons_overview}
\end{table}

\subsubsection{Metrics}

In a biometric system, two types of erroneous decisions can occur, \ie false match or false non-match~\cite{ISO-IEC-19795-1-Framework-210216}: The probability rates of these decision outcomes are expressed as:

\begin{itemize}
    \item \textbf{False Match Rate (FMR)}: The proportion of the completed biometric non-mated comparison trials that result in a false match.
    \item \textbf{False Non-Match Rate (FNMR)}: The proportion of the completed biometric mated comparison trials that result in a false non-match.
\end{itemize}

Following ISO/IEC guidelines, we build upon these metrics and also report  1) detection error tradeoff (DET) curves between FNMR and FMR; 2) the IAPMR, which in~\cite{ISO-IEC-30107-3-PAD-metrics-170227}, for a full system evaluation, is defined as: \textit{the proportion of impostor attack presentations using the same PAI species in which the target reference is matched}. We report the IAPMR at high-security thresholds corresponding to an FMR of 0.1\% and 1\%. \par In the evaluation of DET plots, we consider T-shirt faces compared against digital faces of the same identity as a type of mated comparison.

\subsubsection{Results}
\label{sec:fr_evaluation} 
\begin{figure}[!htb]
\begin{subfigure}[t]{0.48\linewidth}
    \centering
  \includegraphics[width=\linewidth]{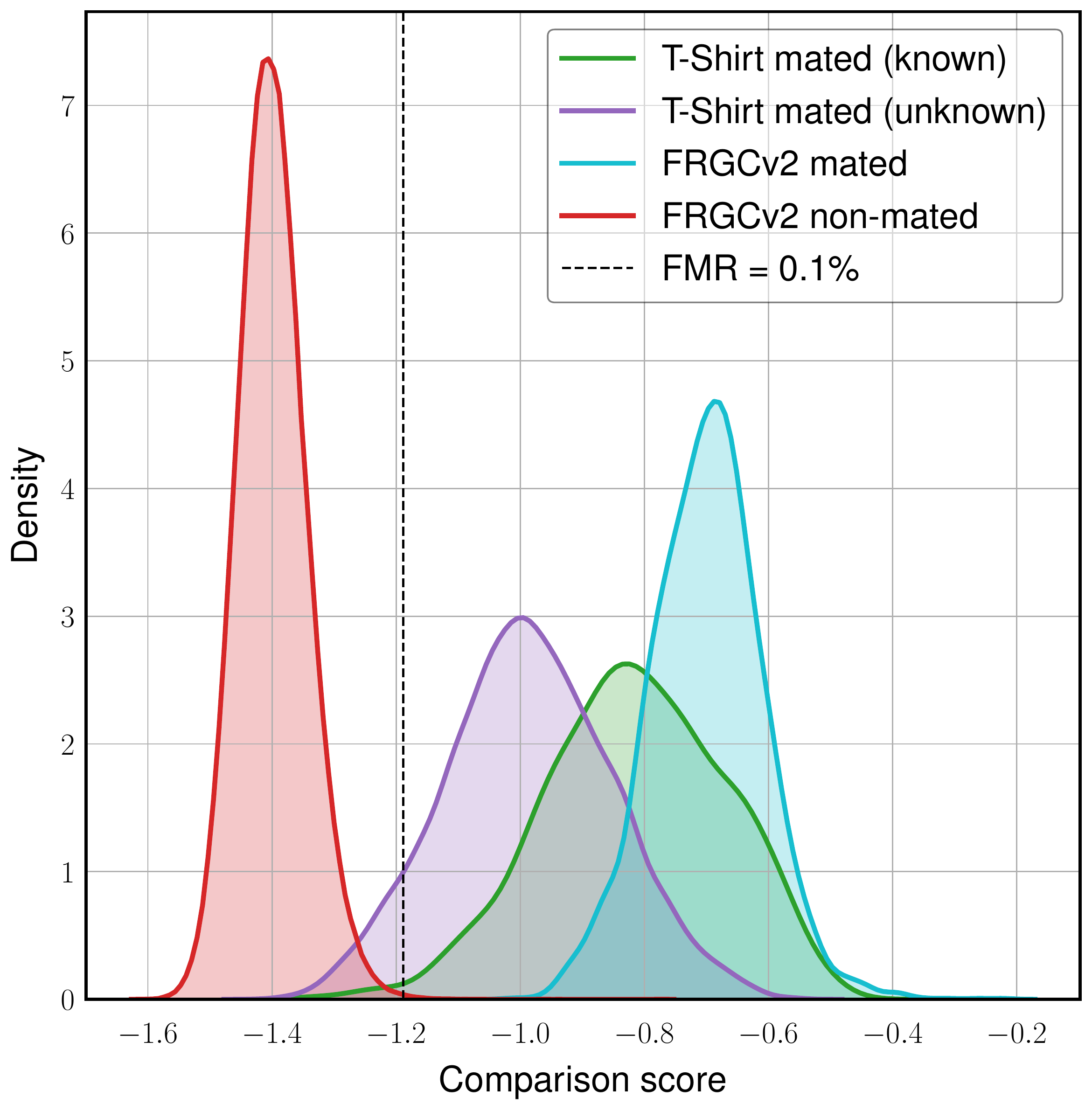}
  \caption{ArcFace}
\end{subfigure}%
\begin{subfigure}[t]{0.48\linewidth}
    \centering
  \includegraphics[width=\linewidth]{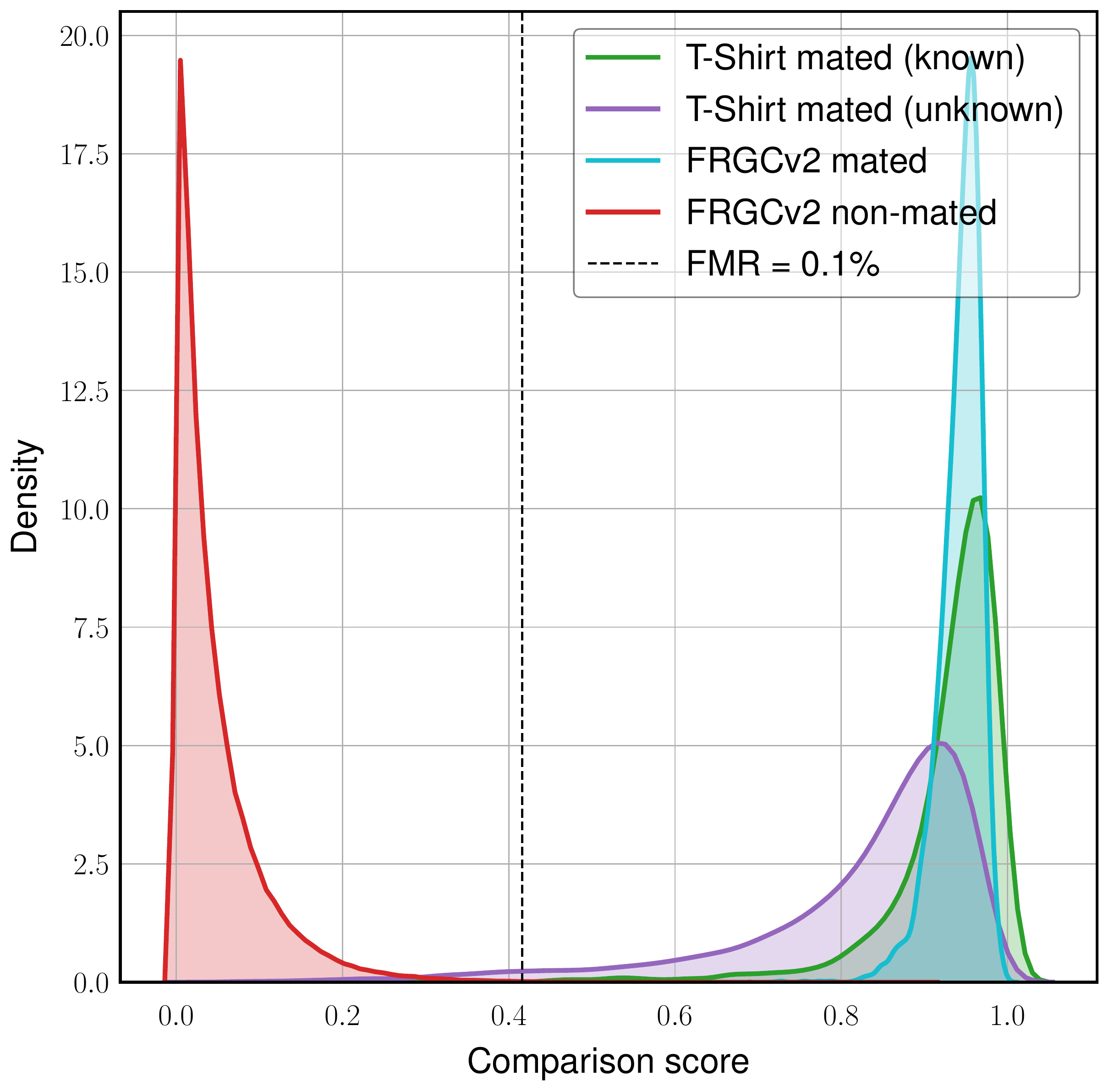}
  \caption{COTS}
\end{subfigure}\quad %
\caption{Face recognition score distributions for PAs and bona fide comparisons.}
\label{fig:tshirt_fr_evaluation_kde}
\end{figure}

\begin{figure}[!htb]
\begin{subfigure}[t]{0.48\linewidth}
    \centering
  \includegraphics[width=\linewidth]{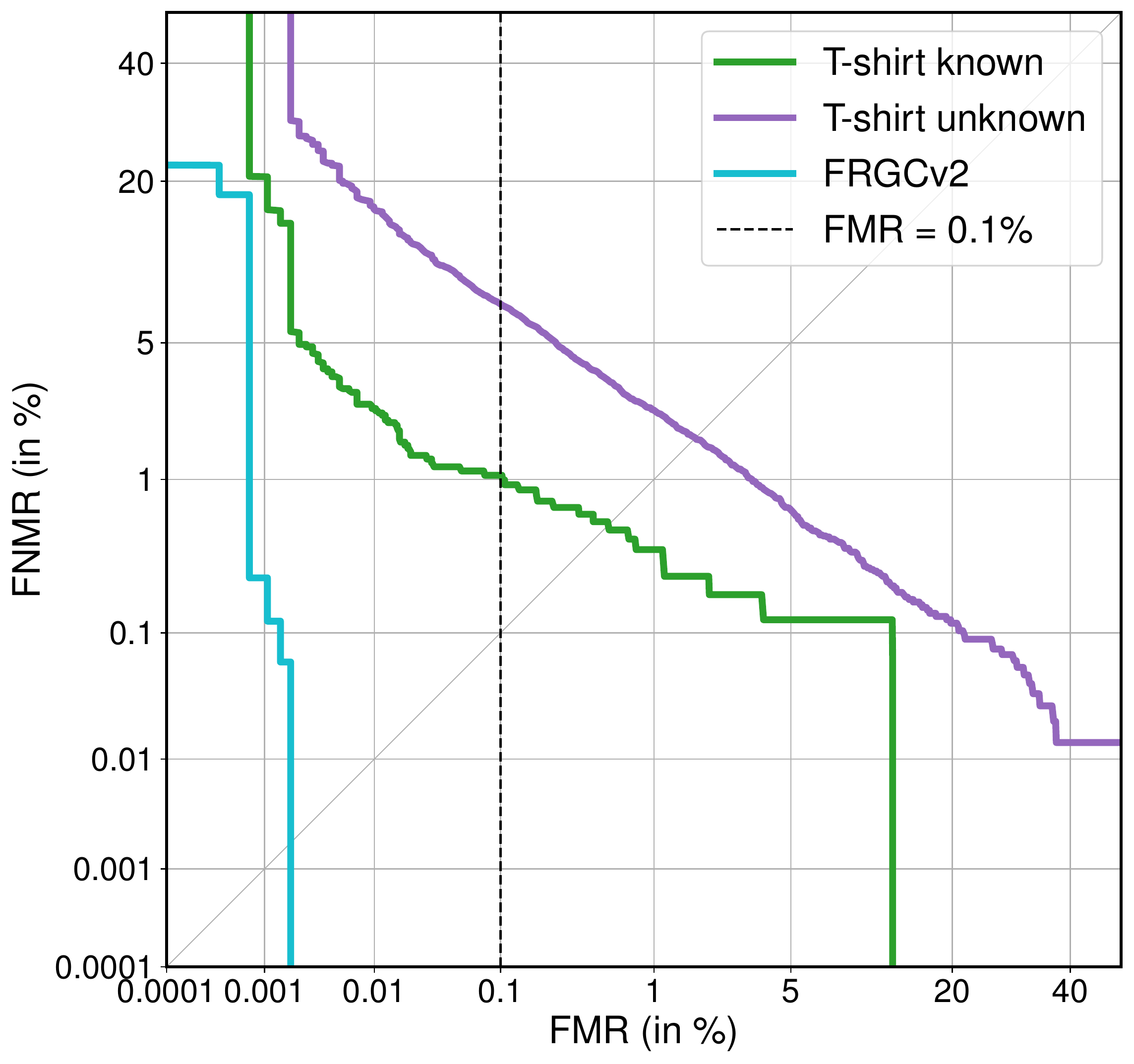}
  \caption{ArcFace}
\end{subfigure}%
\begin{subfigure}[t]{0.48\linewidth}
    \centering
  \includegraphics[width=\linewidth]{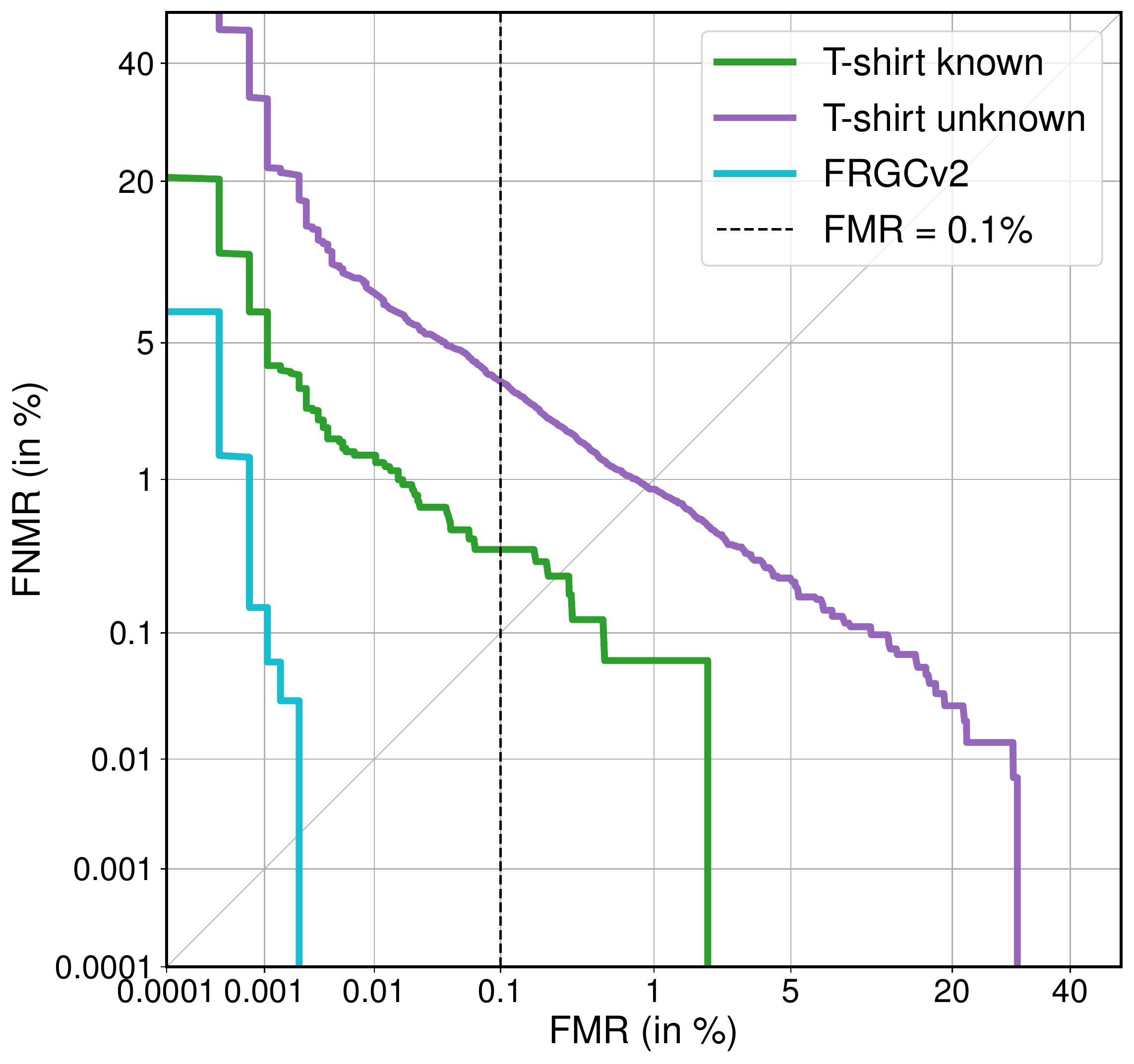}
  \caption{COTS}
\end{subfigure}\quad %
\caption{DET-curves showing recognition performance of the PAs and bona fide images.}
\label{fig:det_plots}
\end{figure}

The score distribution and DET-plots of the T-shirt attacks for the two scenarios and the bona fide images from FRGCv2 are shown in figure~\ref{fig:tshirt_fr_evaluation_kde} and~\ref{fig:det_plots} for both ArcFace and COTS. Additionally, the success rate of the attacks on both face recognition systems is given in table~\ref{tab:iapmr_tshirt_attacks} at two high-security thresholds. The results show a high vulnerability of the two systems, \ie $>$92.6\% IAPMR, for both the known and unknown attack scenarios. 

\begin{table}[h!]
\centering
\caption{IAPMR results calculated at  FMR = 0.1\% and FMR = 0.1\% for both attack scenarios on COTS and ArcFace.}
\label{tab:iapmr_tshirt_attacks}
\begin{adjustbox}{max width=\columnwidth}
\begin{tabular}{@{\extracolsep{2pt}}llll@{}} \toprule 
\multicolumn{1}{c}{}  &  \multicolumn{1}{c}{}  & \multicolumn{2}{c}{\textbf{IAPMR$\%$}} \\ \cmidrule{3-4} 
\multirow{-2}{*}{\textbf{System}} & \multirow{-2}{*}{\textbf{Scenario}}  & \textbf{FMR$=0.1\%$}      & \textbf{FMR$=1\%$}      \\ \midrule
\multirow{2}{*}{ArcFace} & T-shirt known  &  98.9428 &  99.6269       \\
& T-shirt unknown  & 92.6755 & 97.6299         \\ \midrule
\multirow{2}{*}{COTS} & T-shirt known  & 99.6262 & 99.9377       \\
& T-shirt unknown  & 96.6909 & 99.1208          \\ \bottomrule
\end{tabular}
\end{adjustbox}
\end{table}

\section{Presentation Attack Detection}
\label{sec:pad_evaluation}
The purpose of this section is to analyze the capabilities of PAD algorithms for detecting the  T-shirt attacks in an unknown scenario, \ie where T-shirt PAs are not seen during training. To this end, we use two existing state-of-the-art PAD algorithms and propose three new algorithms.

\subsection{Models}
Five algorithms were used to evaluate the capabilities to detect the T-shirt attacks in an unknown scenario, \ie where attacks are not seen during training:

\begin{LaTeXdescription}
  \item[Vision Transformer (VIT)] The VIT approach is based on ~\cite{George-VIT-IJCB-2021} in which the final layer of an existing vision transformer is changed and fine-tuned for binary classification.
\end{LaTeXdescription}

\begin{LaTeXdescription}
  \item[Cross Modal Focal Loss (CMFL)] The CMFL approach is based on~\cite{George-CMFLForRGBDFaceAntiSpoofing-CVPR-2021} in which the authors propose the CMFL loss function to better modulate individual channel contribution when training on multi-channel PAD datasets. A benefit of the CMFL approach is that the network has a two stream multi-head architecture which allows to get the detection scores separately for each input channel and for the joint model. In the results, CMFL (RGBD) will denote the case where the score from the joint architecture is taken using both RGB and depth images, and for CMFL (RGB), only RGB images are used.
\end{LaTeXdescription}

\begin{LaTeXdescription}
  \item[Depth Variance (DV)] The DV approach is a non-learning-based method which analyzes the variance of depth values located at key landmarks. Specifically, given a RGB image and its corresponding depth map in gray-scale, 468 landmarks are found on the RGB image using MediaPipe~\cite{Lugaresi-Mediapipe-CVPR-2019}. Thereafter, the depth-value at each landmark is found and the standard derivation of these values are recorded, resulting in a depth variance score which can be used for PAD.
\end{LaTeXdescription}

\begin{LaTeXdescription}
  \item[Anomaly detection (AD)] The AD approach is a anomaly detection approach trained on a subset of bona fide images from the FRGCv2 subset which includes high quality reference images and more unconstrained probe images captured with varying illumination conditions and alike. As features, a 768-dimensional feature vector is extracted using a pre-trained vision transformer~\cite{Dosovitskiy-VisionTransformer-arxiv-2020, Vit-pretrained-Wightman-github} where the last layer has been removed. The extracted features are given as input to a one-class support vector machine (SVM) with a linear kernel.
\end{LaTeXdescription}

\begin{LaTeXdescription}
  \item[DV + AD (fusion)] This approach performs a score-level fusion of the output of the DV and AD approach using an equal weight factor of 50\%. Prior to score fusion a Min-Max normalization is performed individually on the output of DV and AD on the evaluation data.
\end{LaTeXdescription}

\subsection{Experiments}
To evaluate the capabilities to generalize to the T-shirt attacks, the VIT and CMFL algorithms are trained on the \textit{grandtest} protocol of the \textit{HQ-WMCA} dataset. The dataset comprises 10 unique attacks (see table~\ref{tab:multichannel_pad_databases_overview}) but does not include T-shirt attacks. The anomaly detection approach is trained on a subset of controlled reference images and more uncontrolled probe images from the FRGCv2~\cite{Phillips-FRGC-2005} database. For evaluating the approaches, 1,604 attacks\footnote{four attack images were left out as they could not be properly preprocessed by two of the PAD algorithms} of the TFPA database are used along with 728 bona fide images consisting of faces with a frontal pose. 

\subsection{Metrics}
The metrics used in the evaluation comply with the ISO/IEC 30107-Part 3~\cite{ISO-IEC-30107-3-PAD-metrics-170227} standard for biometric PAD. As a result, we report:

\begin{itemize}
    \item Attack Presentation Classification Error Rate (APCER), which is the proportion of attack presentations misclassified as bona fide presentations.
    \item Bona Fide Presentation Classification Error Rate (BPCER), which is the proportion of bona fide presentations wrongly classified as attack presentations.
\end{itemize}

We report the BPCER observed at different thresholds corresponding to a APCER of 5\% (BPCER20) and 10\% (BPCER10). Additionally, we compute and display DET-curves which shows the trade-off between the APCER and BPCER across different operating points of a PAD system.

\subsection{Results}
\begin{figure}[!htb]
    \centering
    \includegraphics[width=0.8\linewidth]{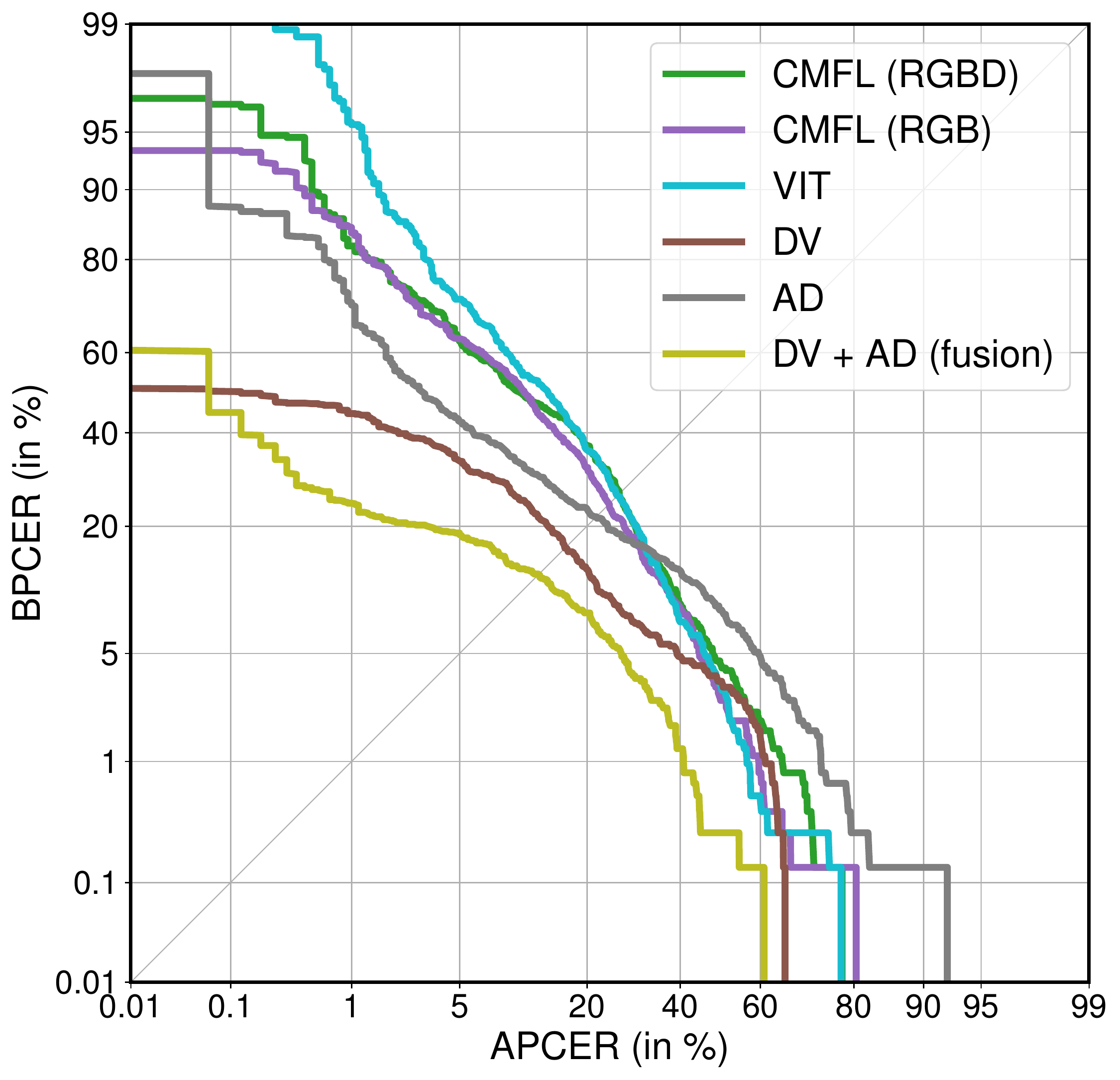}
    \caption{DET curves showing PAD performance on TFPA}
    \label{fig:det_padl}
\end{figure}

The DET curves for the PAD systems can be seen in figure~\ref{fig:det_padl} and relevant performance metrics are shown in table~\ref{tab:pad_metrics_results}. As shown, the best system is the proposed fusion approach which achieves a D-EER of around 12.52\% and BPCER of around 13.46\% and 18.54\% at APCER=10\% and APCER=5\%, respectively. The VIT and CMFL approaches generally achieve the worst performance with D-EER $>$ 25\% whereas the depth variance and anomaly detection approaches perform better with D-EER of 16.34\% and 21.70\%, respectively. While the obtained detection error rates appears to be high at first glance, the results are comparable to current state-of-the-art face PAD performance for unknown PAIs; as mentioned in the beginning.

\begin{table}[!htb]
    \centering
    \caption{D-EER, BPCER10, and BPCER20 in $\%$ for the different PAD algorithms.}
    \begin{tabular}{@{}llll@{}} \toprule \textbf{PAD Algorithm} & \textbf{D-EER} & \textbf{BPCER10} & \textbf{BPCER20} \\ \midrule 
    CMFL (RGBD) & 26.14 & 50.41 & 62.77 \\
    CMFL (RGB) & 24.18 & 52.34 & 63.32   \\
    VIT & 25.82 & 56.59 & 72.25 \\
    DV & 16.34 & 25.55 & 33.52 \\
    AD & 21.70 & 33.24 & 42.72 \\
    DV + AD (fusion) & \textbf{12.52} & \textbf{13.46} & \textbf{18.54} \\
    \bottomrule
    \end{tabular}
    \label{tab:pad_metrics_results}
\end{table}

\section{Discussion}
\label{sec:discussion}
Considering the results, it is clear that the T-shirt attacks pose a potential threat to face recognition systems especially in uncontrolled and semi-controlled applications, \eg in a border control applications and in combination with a stolen passport. In particular, the following observations can be made: 

\begin{LaTeXdescription}
  \item[Face detection] The results show the potential for launching the proposed T-shirt attacks against face recognition systems as it was shown that three state-of-the-art face detection algorithms will almost always and with high confidence detect the faces printed on the T-shirts. It should be noted that if a face detection system simultaneously detects both a real face and a T-shirt face, then the system might potentially fail, which might lead to the system trying to acquire a new sample or request human supervision. For such systems, it might be more difficult to launch the attack. Nevertheless, it might be possible to cover the real face with the hands, as demonstrated in this work, or use other face concealment techniques such as, \eg adversarial attacks, which have shown good results at evading object detection~\cite{Yin-AdversiralAttacksAgainstObjectDetectionAndContextConsistencyChecks-WACV-2022}.
\end{LaTeXdescription}

\begin{LaTeXdescription}
  \item[Face recognition] For face recognition, it was shown that in cases where the T-shirt face is properly detected, it is possible to spoof face recognition systems if no attack detection mechanism or safeguard is implemented into the system. More specifically, we showed that it is possible to spoof the face recognition system with high success rate, \ie IAPMR $>$ 92.6\%, for the tested open-source and commercial face recognition systems. The results also show the highest vulnerability for the commercial system, indicating that systems which perform better on bona fide data are more vulnerable to PAs; this is in line with previous findings (see section~\ref{sec:related_work}).
\end{LaTeXdescription}

\begin{LaTeXdescription}
  \item[Presentation attack detection] The results for PAD show limited generalizability of two open-source algorithms when trained on the \textit{grandtest} protocol of HQ-WMCA. Interestingly, from the PAD results for the CMFL approach, we saw that using depth information does not yield a better performance than just using RGB information despite processing the data in the same way for the two datasets. These observations are likely due to variations in how the depth data was acquired across the two datasets such as difference in distance from camera to the subject. Cross-domain generalizability is a known challenge to PAD algorithms which needs to be further investigated and addressed in the future. \par From just analyzing statistical differences between the depth maps of real faces and faces on the T-shirts, we saw that a analysis of the depth maps makes it possible to distinguish between T-shirt attacks and real faces to some degree. This observation indicates that it is possible to use the depth stream to accurately detect the T-shirt attacks using just the depth data present in the face region. The proposed anomaly detection approach also showed promising results using only RGB data. Additionally, fusion of the anomaly detection approach and depth variance approach showed the best results. \par In this work all the PAD algorithms were evaluated after the preprocessing stage of a face recognition, \ie where faces are already detected and cropped. However, it is likely that T-shirt PAs will have a reduced attack potential if areas outside the face region are processed, \eg by learning the location of the face relative to other body parts such as the shoulders. Such an investigation was deliberately left out of this paper as it would make a comparison to existing approaches unfair as these mostly consider areas within the face region.
\end{LaTeXdescription}

\section{Conclusion}
\label{sec:conclusion}
Face recognition has become more prevalent with advancements in deep learning and the availability of large face databases. Despite this, it is clear that such systems are vulnerable to presentation attacks and must be accompanied by robust methods for detecting such attacks, which otherwise can be used to circumvent the security of a face recognition system. Recently, methods for detecting presentation attacks have moved towards generalizable presentation attack detection methods capable of detecting many different types of attacks, even attacks which are unseen during training. To accurately measure the capabilities of such presentation attack detection methods and advance the field, benchmarks which contain new and novel attack types must be introduced. This work introduced a novel multi-channel T-shirt Face Presentation Attack (TFPA) database containing 100 unique T-shirt PAIs and 1,608 T-shirt impersonation attacks. The vulnerability of face recognition systems to the attacks was evaluated by first analyzing the feasibility of launching the attacks and subsequently evaluating the success rate of the T-shirt attacks. The results showed that it was possible to launch the attack if the real face was covered. In such cases where the face on the T-shirt is detected, the attack success rate was higher than 92.6\% in terms of IAPMR for both an open-source and a commercial face recognition system. Additionally, three new presentation attack detection algorithms were proposed and used along with two existing state-of-the-art algorithms to measure the detectability of T-shirt impersonation attacks in an unknown scenario where T-shirt attacks were not seen during training. Best results were achieved by the proposed score level fusion of an anomaly detection approach trained on bona fide RGB images and a statistical detection model which analyses the variance of depth values measured at key landmarks of a face. The approach obtained a D-EER of 12.52\% on bona fide and T-shirt attacks from the TFPA database. By making the collected database available, research groups are encouraged to improve upon the detection results obtained in this work and prevent from T-shirt attacks in the future. 

\section{Acknowledgement}
This research work has been partially funded by the German Federal Ministry of Education and Research and the Hessian Ministry of Higher Education, Research, Science and the Arts within their joint support of the National Research Center for Applied Cybersecurity ATHENE, the European Union's Horizon 2020 research and innovation programme under the Marie Sk\l{}odowska-Curie grant agreement No. 860813 - TReSPAsS-ETN.  

\bibliographystyle{IEEEtran}
\bibliography{bibliography}

\end{document}